\documentclass[11pt]{article}

\usepackage[preprint]{acl}

\usepackage{times}
\usepackage{latexsym}

\usepackage[T1]{fontenc}

\usepackage[utf8]{inputenc}

\usepackage{microtype}

\usepackage{inconsolata}

\usepackage{graphicx}

\usepackage{amsmath}
\usepackage{amsthm}
\usepackage{booktabs}
\usepackage{algorithm}
\usepackage{algorithmic}
\usepackage[switch]{lineno}

\newcommand*\circled[1]{
    \raisebox{.5pt}{\textcircled{\raisebox{-.9pt} {#1}}}
}
\usepackage{pifont}
\usepackage{graphicx}
\usepackage{booktabs}
\usepackage{subcaption}
\usepackage{enumitem}
\usepackage[table]{xcolor}
\usepackage{natbib}
\usepackage{multirow}

\title{Do not be greedy, \textsc{Think Twice}:\\Sampling and Selection for Document-level Information Extraction}

\author{
Mikel Zubillaga \quad Oscar Sainz \quad Oier Lopez de Lacalle \quad Eneko Agirre\\
HiTZ Center - Ixa, University of the Basque Country EHU\\
\texttt{mikel.zubillaga@ehu.eus}
}

\begin{document}

\maketitle

\begin{abstract}
Document-level Information Extraction (DocIE) aims to produce an output template with the entities, relations, and events of interest occurring in the given document. Standard practices include prompting decoder-only LLMs using greedy decoding to avoid output variability. Rather than treating this variability as a limitation, we show that sampling can produce substantially better solutions than greedy decoding, especially when using reasoning models. We thus propose \textsc{ThinkTwice}, a sampling and selection framework in which the LLM generates multiple candidate templates for a given document, and a selection module chooses the most suitable one. We introduce both an unsupervised method that exploits agreement across generated outputs, and a supervised selection method using reward models trained on labeled DocIE data. To address the scarcity of golden reasoning trajectories for DocIE, we propose a rejection-sampling-based method to generate silver training data that pairs output templates with reasoning traces. Our experiments show the validity of unsupervised and supervised \textsc{ThinkTwice}, consistently outperforming greedy baselines and the supervised state-of-the-art. Our code is publicly available.\footnote{\href{https://github.com/MikelZubi/ThinkTwice}{https://github.com/MikelZubi/ThinkTwice}}
\end{abstract}

\begin{figure}[t]
    \centering
    \includegraphics[width=\linewidth]{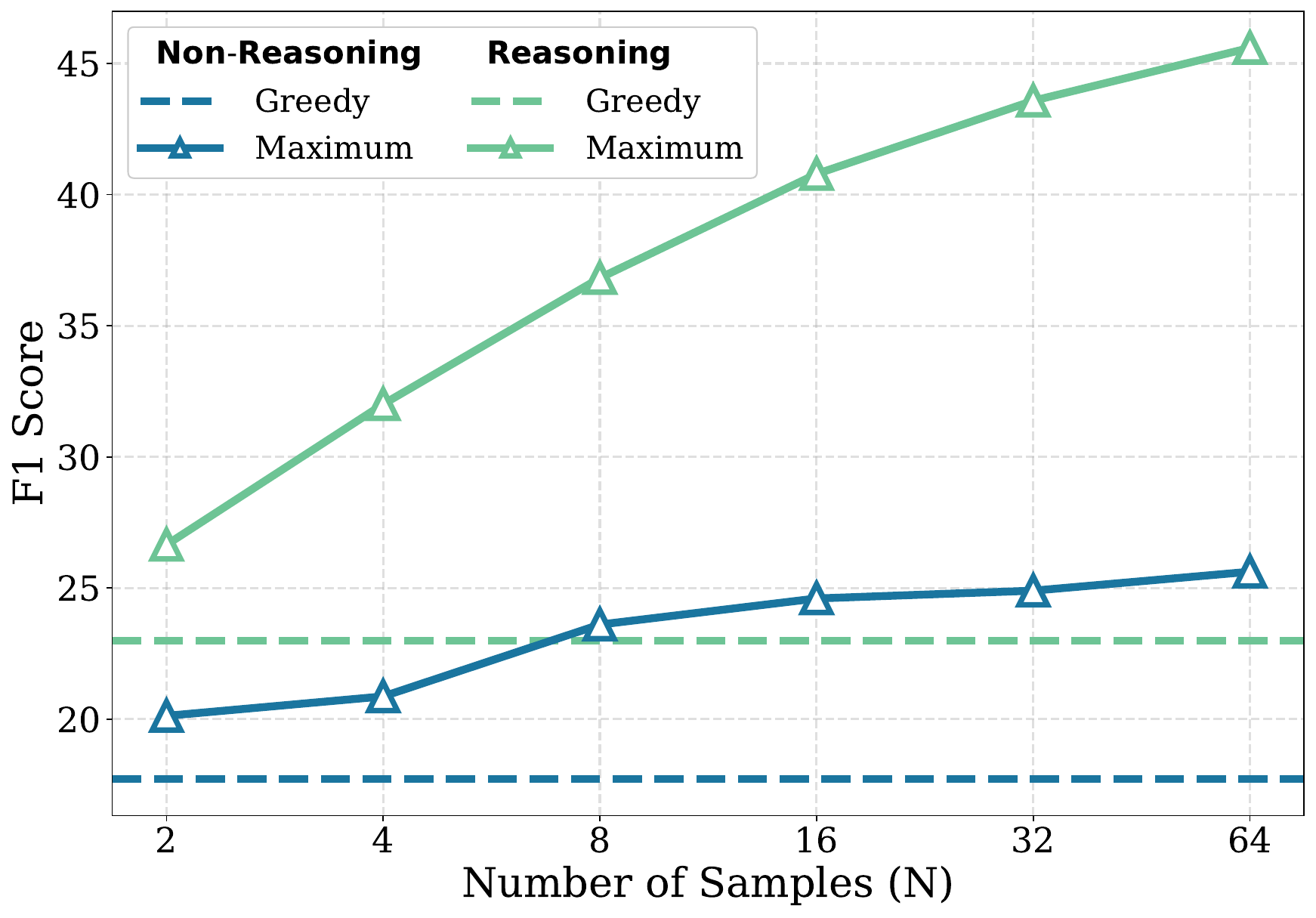}
    \caption{Sampling results on MUC-4, \emph{maximum} reports the results of oracle selection among generated samples.} 
    \label{fig:variability}
\end{figure}

\section{Introduction}

Information Extraction (IE) aims to convert naturally occurring text into structured representations by identifying entities, attributes, and relations of interest. Traditionally, IE was conceived as a document-level task~\citep{muc-1992-message, ji2010overview}, but with the popularization of statistical and machine-learning methods, the field shifted towards sentence-level datasets~\citep{ACE,zhang2017tacred,Grishman_2019}, which were simpler and easier to annotate. Still, many real-world scenarios require reasoning over entire documents, where relevant information is scattered across multiple sentences or sections. Document-level IE (DocIE) addresses this challenge by jointly extracting and aggregating information at the document scope, typically producing structured templates that depend on long-range context and cross-sentence interactions. With the rapid development of Large Language Models (LLMs), and their longer context, it has now become more natural to work on the document-level~\citep{gatto-etal-2025-document} instead of analyzing each sentence independently and aggregating the results~\citep{chen-etal-2023-iterative}.

In contrast to encoder-based models, decoder-only LLMs can be directly prompted with some annotation guidelines to produce the annotations for a given document, enabling zero- and few-shot inferences without task-specific architectural modifications or fine-tuning. The generative nature of LLMs, however, presents a challenge when it comes to evaluating their performance, as each time they are prompted, a different solution could be generated. In order to produce reproducible comparisons, researchers leverage greedy decoding strategies to ensure the same output template is produced for a given pair of guidelines and document~\citep{GoLLIE2024sainz,li-etal-2023-revisiting-large}. Producing only a single answer---even though the most probable is chosen at each time---may not produce the best answer, hindering the evaluation and hiding the real performance of a given model. As we show in Figure~\ref{fig:variability}, there exists an increasing gap between the best-performing answer and the greedy-decoded answer when the number of samples increases, which further intensifies with reasoning models. This phenomenon is not tailored to IE only, but recently studied under the Test-Time Scaling (TTS) paradigm, particularly for mathematical reasoning tasks~\citep{wu2025inference,snell2025scaling}.

The application of TTS techniques to IE remains largely unexplored. To address this gap, we propose \textsc{ThinkTwice}, a sampling and selection framework in which an LLM generates multiple candidate extractions for a document, and a dedicated selection function identifies the most reliable output. We show that greedy decoding fails to fully exploit the capabilities of LLMs, motivating the use of sampling-based decoding strategies. We further analyze how this effect manifests in both reasoning-oriented and non-reasoning models.

As our main \textbf{contributions}, we introduce a generation and selection framework, as well as two new selection strategies tailored to IE: an unsupervised \textbf{F1 Voting} method and a supervised \textbf{reward-based} selection strategy. Both substantially outperform standard TTS baselines and establish new supervised state-of-the-art results in document-level IE. In the supervised setting, we overcome the lack of golden trajectories for DocIE with a rejection-sampling approach for generating high-quality reasoning traces, which are used to produce silver training data. Our experimental results lead to the following key \textbf{findings}: 1) Reasoning models consistently outperform standard LLMs in document-level IE. 2) The proposed unsupervised sampling and selection framework yields consistent gains over the greedy decoding and standard TTS baselines in zero-shot settings. 3) Training on high-quality silver reasoning trajectories paired with output templates achieves state-of-the-art results in document-level IE when combined with a supervised selection strategy, both in the monolingual and cross-lingual settings.

\section{Related Work}

\paragraph{Document-level Information Extraction.} Information Extraction, while originally conceived as a document-level task, was long constrained to the sentence level due to the limitations of earlier systems. With recent advances in the field of Natural Language Processing (NLP), however, DocIE has regained significant attention, on problems such as argument linking \citep{li-etal-2021-document}, \textit{N}-ary relation extraction \citep{jain-etal-2020-scirex}, and, our primary focus, template extraction \citep{huang-etal-2021-document,chen-etal-2023-iterative}. This resurgence has led researchers to revisit classic datasets such as MUC \citep{grishman-sundheim-1996-message} and to develop new resources like BETTER \citep{mckinnon-rubino-2022-iarpa} and the multilingual extension of MUC-4, MultiMUC \citep{gantt-etal-2024-multimuc}. The state-of-the-art on these datasets is \textsc{IterX}~\citep{chen-etal-2023-iterative}, which formulates template extraction as a Markov decision process, generating templates iteratively using an imitation-learned policy, and introduces a new evaluation metric for MUC-4, CEAF-RME. However, \textsc{IterX} relies on encoder-based backbones, which are unsuitable for zero-shot settings and often require truncating the document due to their shorter sequence length, resulting in lower performance. Motivated by these limitations, recent work has begun to explore the use of LLMs for IE.

\paragraph{LLMs for Information Extraction.} The adoption of large language models in IE has grown substantially in recent years \citep{zhang-etal-2025-survey}. In addition to achieving state-of-the-art performance in supervised settings \citep{wadhwa-etal-2023-revisiting,ding-etal-2024-rethinking}, LLMs have demonstrated strong zero-shot IE capabilities by leveraging dataset-specific guidelines \citep{GoLLIE2024sainz}. This flexibility extends the Unified Information Extraction paradigm and opens new research directions. LLMs can also use in-context learning examples to improve their performance without further training \citep{wan-etal-2023-gpt,zhang2024incontextlearningfewshotnested} or be decoded using constrained decoding to create always valid outputs \citep{sakota-west-2025-combining}. However, the inherently stochastic nature of LLMs leads to variability across generations. Although greedy decoding is commonly used to reduce this instability, our results suggest that it is not the most effective decoding strategy. Motivated by this observation, we explore a sampling and selection approach inspired by recent test-time scaling methods.

\paragraph{Reasoning and Test-Time Scaling.} With the release of OpenAI’s o1 model~\citep{openai2024openaio1card}, reasoning-focused models have become a central topic of interest in the field. At the same time, Test-Time Scaling (TTS) has also gained attention \citep{zhang2025surveytesttimescalinglarge}. This technique increases computational effort during inference to achieve better results. To support this approach, several methods have emerged, such as Majority Voting ~\citep{wang2023selfconsistency,chen2024are}, which selects the most frequent answer among multiple sampled responses, and the use of reward models~\citep{wu2025inference,lightman2023letsverifystepstep} to identify the highest-quality output. However, most existing work in this area has primarily focused on math and reasoning-intensive benchmarks \citep{zhang2025surveytesttimescalinglarge}. The methodology proposed in this work draws inspiration from these TTS approaches while adapting them to the IE task.

\section{Methodology}

\begin{figure*}[t!]
    \centering
    \includegraphics[width=1.0\linewidth]{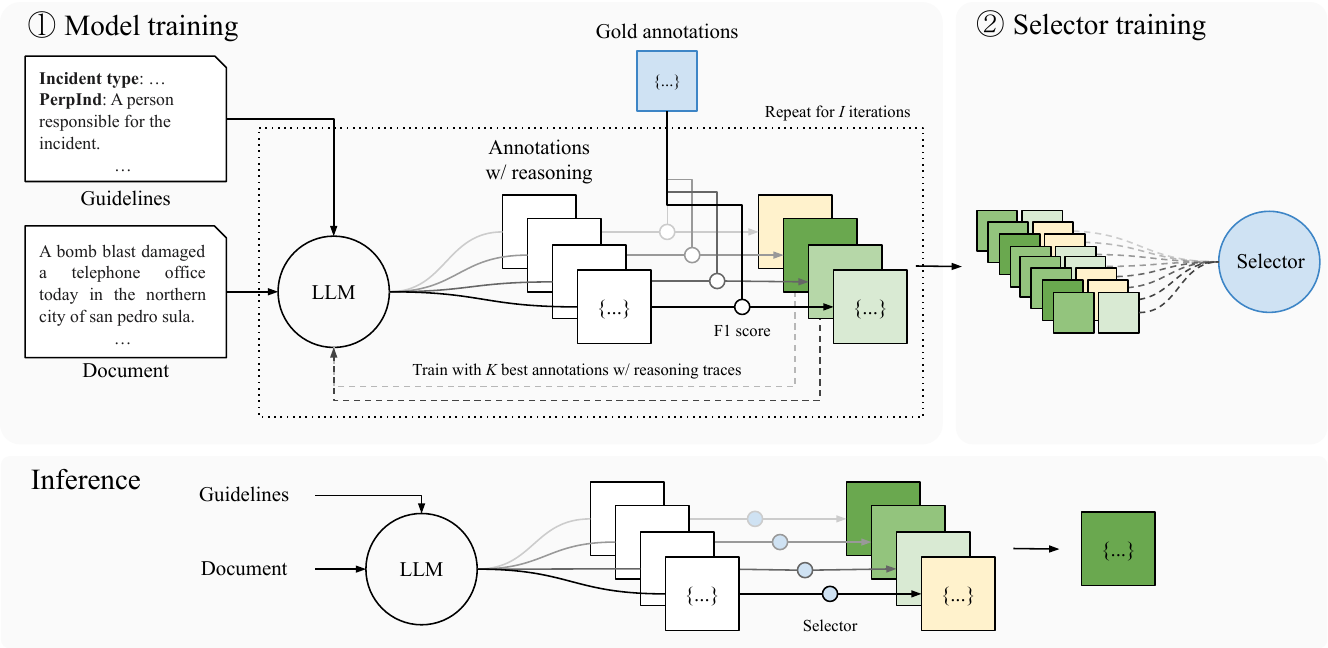}
    \caption{\textsc{ThinkTwice} architecture, with the inference process at the bottom. The supervised option includes two steps: \circled{1} The iterative procedure to generate the silver dataset with trajectories and to fine-tune the reasoning model; \circled{2} Training the selector with silver preference data that include trajectories.}
    \label{fig:method}
\end{figure*}

While standard approaches rely on greedy decoding to produce a single output, Figure \ref{fig:variability} shows that the highest-scoring outputs are not always obtained with this method. Generating multiple candidate templates and applying a selection strategy could potentially yield substantially better results. Moreover, the gap between greedy decoding performance and the maximum achievable score \textbf{is even more pronounced for reasoning models}. Training reasoning models, however, requires good-quality reasoning traces along with the gold standard templates, which are not available for our current datasets. 

To address these issues, we propose \textsc{ThinkTwice}, a sampling and selection architecture, where the model first produces several outputs for a given input, and the selector then decides the final output. In addition, in order to properly train a model based on reasoning LLMs, we propose a rejection-sampling-based method to generate good-quality reasoning traces for each training sample.

\subsection{Task Representation}

Following previous works on zero-shot IE, we adopted the annotation guidelines aware prompting strategy~\citep{GoLLIE2024sainz} as our task representation. Formally, for a given document-level IE dataset $\mathcal{D}$, each example is a triplet $(g, x_i, t_i)$ where $g$ are the annotation guidelines for the given dataset, $x_i$ is the text (or document) to analyze, and, $t_i$ is the template with the structured information associated with $x_i$ text. Current LLMs require following a predefined chat template that specifies how each message in a conversation is encoded and identifies the author role of each message (i.e., system, user, or assistant). The final instance representation is obtained by giving the annotation guidelines $g$ encoded as Markdown text within the system prompt, and each text document $x_i$ in the user prompt. The output template $t_i$ is expected to be generated by the model under the assistant role following a predefined JSON schema. More information about the prompts we used can be found in Appendix \ref{app:prompt}.

\subsection{Inference}

The inference of our method is described in Figure~\ref{fig:method}. Briefly, \textsc{ThinkTwice} consists of performing a two-step inference where, first, a model $\pi_\theta$ is prompted with task guidelines $g$ and a document $x_i$ to produce $N$ possible template candidates $T_i = \{\hat{t}_{i,1}, \ldots, \hat{t}_{i,N}\}$. On a second step, a selector strategy $\mathcal{S}(g, x_i, T_i)$ is applied over the generated options to select the preferred output candidate. This selector $\mathcal{S}$ can be either an unsupervised or supervised method, as described later in Section~\ref{subsec:selection_strategies}.

Producing complex document-level IE output templates can be challenging, particularly in zero-shot scenarios. To that end, we sampled the candidates applying a constrained decoding strategy where the model was forced to follow a predefined JSON schema. In the case of reasoning models, we first let the model generate the reasoning traces, and then apply the same constrained decoding strategy to force the output to meet the requirements. The JSON schemas leveraged during the inference are described in Appendix \ref{app:json_schemas}.

\subsection{Model training}\label{subsec:training}

Training standard LLMs for the task can be accomplished by simply optimizing the cross-entropy loss for each training example $(g, x_i, t_i)$ in the dataset $\mathcal{D}$. However, to fine-tune a reasoning LLM, we also require good-quality reasoning traces $r_i$ tailored to each training example. To that end, we propose a rejection-sampling-based method to generate those reasoning traces that we require to train our model. Figure~\ref{fig:method} illustrates the overall workflow of the proposed iterative approach.

For each input $x_i \in \mathcal{D}$, we first generate $N$ synthetic reasoning–template pairs $(\hat{r}_{i,j}, \hat{t}_{i,j})$ by sampling a reasoning model $\pi_\theta$ conditioned on the instance $(g, x_i)$. We then rank these candidates using a scoring function $\phi(\hat{t}_{i,j}, t_i)$ based on the evaluation metric of the dataset (cf. Section \ref{subsec:evalationdatasets}), which compares each synthesized template to the gold annotation $t_i$, and retain the top $K$ examples while discarding the remaining $N-K$. Using this ranking, we construct a silver dataset $\hat{\mathcal{D}} = \{ (g, x_1, \hat{r}_{1,1}, \hat{t}_{1,1}), \ldots, (g, x_{|\mathcal{D}|}, \hat{r}_{|\mathcal{D}|,K}, \hat{t}_{|\mathcal{D}|,K})\}$, which is subsequently used to train the LLM. This iterative procedure is repeated until the maximum selection score no longer improves across successive iterations. After convergence, we perform a final aggregation step that collects candidates from the preceding iteration, selects the top $R$ reasoning–template pairs, and fine-tunes a newly reinitialized model on this resulting dataset.

\subsection{Selectors and their training}\label{subsec:selection_strategies}
Selecting a single template $\hat{t}_{i,j}$ from a set of candidate templates $T_i = \{\hat{t}_{i,1}, \ldots, \hat{t}_{i,N}\}$ can be approached in multiple ways, either unsupervised or supervised. Standard TTS approaches typically rely on \textbf{Majority Voting}~\citep{wang2023selfconsistency,chen2024are} to select the most frequent candidate. However, while effective for math-related tasks, this strategy often fails for IE, where complex outputs lead to the most common prediction being empty. To address this limitation, we \textbf{propose both an unsupervised and supervised selector}. 

\paragraph{F1 Voting.}
This is an alternative to Majority Voting, specifically designed for IE, which uses F1-score-related metrics\footnote{See Section~\ref{subsec:evalationdatasets} for the metric used in each dataset.} as a continuous similarity measure instead of the binary match signal used in Majority Voting. Each candidate template is assigned a score based on its similarity to the remaining candidates. The candidate with the highest aggregate score is selected. Formally,

\begin{equation}
    \hat{t}_i
    =
    \arg\max_{\hat{t}_{i,j} \in T_i}
    \sum_{k=1}^{N}
    \phi\!\left(\hat{t}_{i,j}, \hat{t}_{i,k}\right),
    \label{eq:sum_f1voting}
\end{equation}

\noindent
where $\phi$ denotes the scoring function (the F1-score in our case), $T_i$ is the set of templates generated for instance~$i$, and $N = |T_i|$.

\paragraph{Supervised selector (reward model).}
Training a reward model requires outputs of varying quality, thus, we propose to train the selector based on the silver instances generated for model training, including poor-quality outputs, together with the gold annotations. As shown in Figure~\ref{fig:method}, the discarded noisy templates produced during training can be leveraged together with the selected examples to learn a preference-based reward model, which is subsequently used to rank candidate templates and select the most suitable one. Formally, we can construct a preference dataset $\mathcal{P}$ based on silver $T_i = \{\hat{t}_{i,1}, \ldots, \hat{t}_{i,N}\}$ and gold $\{t_i\}$ annotations.

\begin{equation}
\begin{aligned}
\mathcal{P}
=
\bigl\{
(g, x_i, t^c_i, t^r_i)
\;\big|\;&
t^c_i, t^r_i \in T_i \cup \{t_i\}, \\
& \phi(t^c_i, t_i) > \phi(t^r_i, t_i)
\bigr\}
\end{aligned}
\end{equation}

\noindent where $t_i^c$ and $t_i^r$ represent the selected and rejected solutions, respectively. Additionally, we can compute an informed margin $m$ by considering the difference in the score between $t_i^c$ and $t_i^r$:

\begin{equation}
m = \lambda \cdot(\phi(t^c_i,t_i) - \phi(t^r_i,t_i))
\end{equation}

\noindent and, $\lambda$ is a scaling factor which we set to 3 as suggested by ~\citet{touvron2023llama2openfoundation}. We can then train the reward model $r_\theta$ by optimizing the Bradley-Terry formula with a margin loss function~\citep{BradleyTerry1952, touvron2023llama2openfoundation}:

\begin{equation}
\resizebox{\columnwidth}{!}{$
\displaystyle
\mathcal{L}_{\mathrm{BT}}
=
\!\!\!\!\!\!\!\!\!\!\!\!\!\!
\sum_{\substack{(g, x_i, t^c_i, t^r_i, m_i) \in \mathcal{P}}}
\!\!\!\!\!\!\!\!\!\!\!\!\!\!\!
-\log \Bigl(
\sigma\bigl(
r_\theta(g, x_i, t^c_i) - r_\theta(g, x_i, t^r_i) - m_i
\bigr)
\Bigr)
$}
\end{equation}

\noindent Finally, we can use the trained reward model $r_\theta$ to rank and select the correct template $\hat{t}_i$ among the candidates as follows:

\begin{equation}
    \hat{t}_i
    =
    \arg\max_{\hat{t}_{i,j} \in T_i}
    r_\theta\!\left(g, x_i,\hat{t}_{i,j}\right),
    \label{eq:infer_reward}
\end{equation}

\section{Experimental Setup}

\subsection{Models}
We leveraged two open-source model families as our backbone models: Llama and Qwen. In addition, we used an embedding model as initialization for our supervised selector approach.

\paragraph{Llama~3.3~70B Instruct} \citep{grattafiori2024llama3herdmodels} is a well-established open-source LLM that achieves strong performance across a wide range of tasks. We use this model as our non-reasoning Llama-based LLM backbone.

\paragraph{DeepSeek Distill Llama~R1~70B} is a distilled version of the DeepSeek~R1~671B model, trained using Llama~3.3~70B as base ~\citep{deepseekai2025deepseekr1incentivizingreasoningcapability}. Since no official reasoning-oriented models have been released in the Llama family, we adopt this model as our Llama-based reasoning backbone.

\paragraph{Qwen3~32B} is the largest dense model in the Qwen3 family and supports both non-reasoning (\textit{no-think}) and reasoning (\textit{think}) inference modes~\cite{yang2025qwen3technicalreport}. This model is particularly well suited to our experiments, as it allows inference with and without reasoning traces while achieving performance comparable to Llama~3.3~70B.

\paragraph{Qwen3~Embedding~8B} \citep{qwen3embedding} is a pretrained embedding model that achieves state-of-the-art performance across a wide range of downstream tasks. We build our reward model on top of this backbone, leveraging its rich semantic representations for reward modeling. To adapt the model to our setting, we add a feed-forward projection head that maps the embedding representation to a single scalar value, which is used as the reward score during inference.

For all experiments, we fix the number of sampled generations per instance to $\mathbf{N = 64}$, which offers a favorable trade-off between exploration and computational cost. Additional hyperparameter details are provided in Appendix~\ref{app:hyperparameters}. Regarding the reasoning token budget, we set a maximum limit of 4000 tokens. In Appendix \ref{app:reasoning}, we analyze the number of reasoning tokens used by each model across the different datasets and languages.

To evaluate the stability of our methods and improve the reliability of the reported results, we perform bootstrapping using three different random seeds over the sampled generations. We report both the mean and standard deviation across these runs.

\subsection{Evaluation Datasets}

\label{subsec:evalationdatasets}

\paragraph{MUC-4 and MultiMUC:} the MUC-4 dataset \citep{muc-1992-message} contains 1{,}700 English news articles on geopolitical conflicts, annotated with structured templates covering six event types: Attack, Arson, Bombing, Kidnapping, Robbery, and Forced Work Stoppage. MUC-4 has served as a foundational benchmark for template-based event extraction. Regarding the evaluation metrics, we report results using the \textbf{CEAF-RME} metric proposed by~\citet{chen-etal-2023-iterative}. Building on this benchmark, MultiMUC \citep{gantt-etal-2024-multimuc} extends the dataset to a multilingual setting by providing parallel annotations in Arabic, Farsi, Korean, Russian, and Chinese alongside the original English corpus.

\paragraph{BETTER Granular} \citep{mckinnon-rubino-2022-iarpa} dataset further increases task difficulty by covering a broader range of domains and event types, including protests, epidemics, natural disasters, acts of terrorism, corruption incidents, and human migration. While it also focuses on six complex event categories, BETTER Granular features substantially longer documents and more intricate templates with finer-grained argument structures, being much more complex than the MUC-4 dataset. Due to the length of the documents and annotation guidelines, we used BETTER Granular exclusively for zero-shot evaluations. While we used the official scorer to compute the final results, we implemented a simplified F1 scorer in order to use it with the F1 Voting selector.

\subsection{Baselines}

We compared \textsc{ThinkTwice} against our baseline, \textbf{greedy decoding}, and previous zero-shot and supervised state-of-the-art methods:

    \paragraph{Zero-shot methods.} Prior work on these datasets under zero- or few-shot settings is limited. Therefore, we evaluate and report \textbf{GPT\textsubscript{5.5}} \footnote{2026/04/23 snapshot.} results in a zero-shot setting on the MUC-4, MultiMUC, and BETTER datasets for reference. 
    
    \paragraph{Supervised methods.} Our main points of comparison for the supervised setting come from the results reported by~\citet{chen-etal-2023-iterative}. We compare our system against \textsc{IterX}~\citep{chen-etal-2023-iterative}, a T5\textsubscript{large enc}-based system and current state-of-the-art; \textsc{GTT}~\citep{du-etal-2021-template}, a BERT encoder based approach; and, \textsc{TempGen}~\citep{huang-etal-2021-document}, a BART encoder-decoder based approach.

\section{Results}

\subsection{To think, or not to think?} 

\begin{figure}[t]
    \centering
    \includegraphics[width=\linewidth]{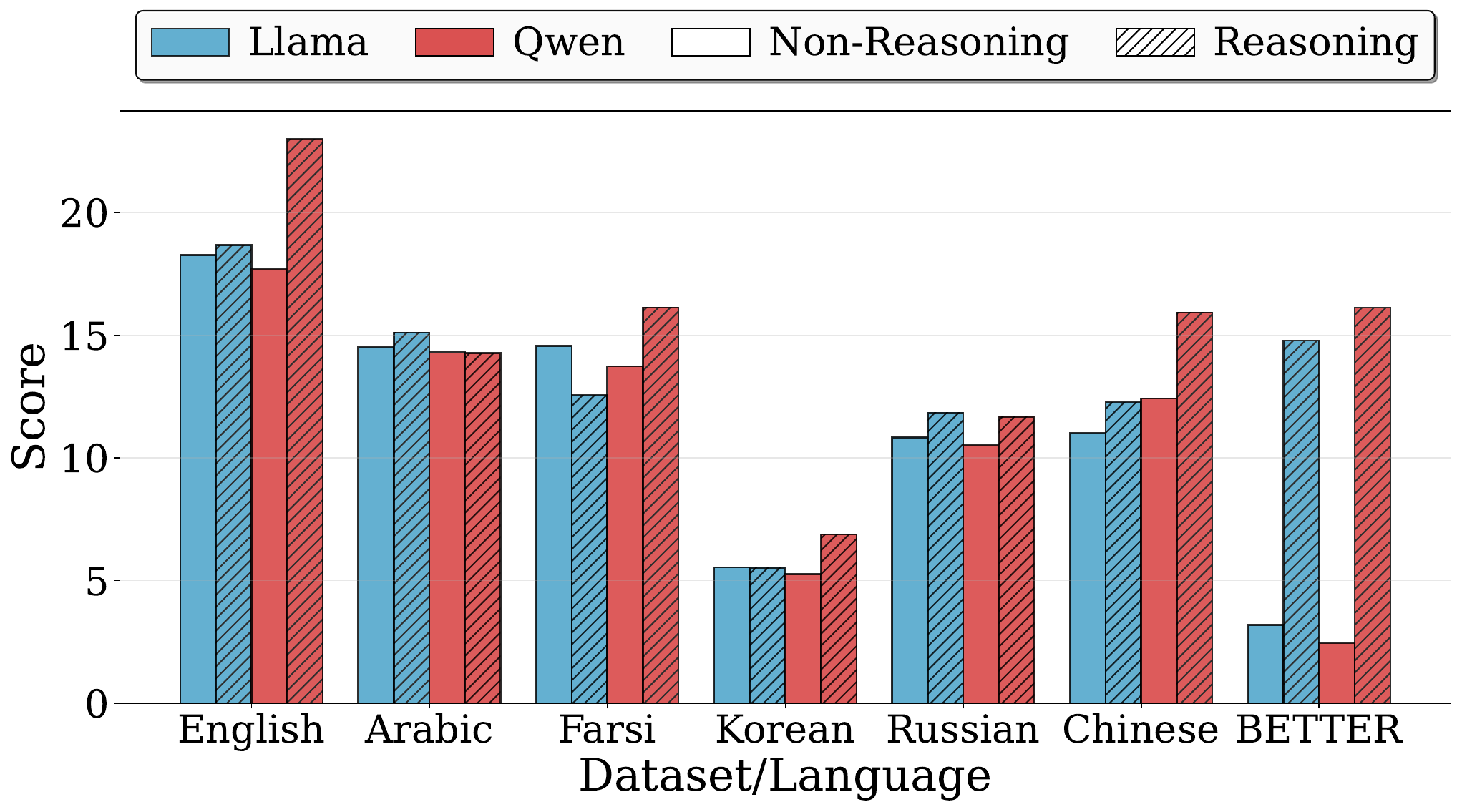}
    \caption{Zero-shot greedy decoding results for the reasoning and non-reasoning baselines of two LLMs on MUC (English), MultiMUC and BETTER.
    }
    \label{fig:think_vs_nothink}
\end{figure}

As depicted by the preliminary results shown in Figure~\ref{fig:variability}, the reasoning models not only perform better, but also allow for bigger \textit{maximum scores}\footnote{We compute the maximum attainable score by selecting the best generated output for each instance, which serves as an oracle upper bound for the selector.}. To properly answer this question, we evaluated both Llama and Qwen models with and without reasoning on all of our testing datasets. Figure~\ref{fig:think_vs_nothink} shows the \textbf{zero-shot} performance of each model-inference combination on each specific dataset. We leveraged \textit{greedy decoding}, i.e., we did not use any sampling nor selector on this initial analysis to focus only on the reasoning/non-reasoning comparison.

\paragraph{Do reasoning models perform better than their non-reasoning counterparts?} As shown in Figure~\ref{fig:think_vs_nothink}, reasoning models perform consistently better than their non-reasoning counterparts, with only a few exceptions. Interestingly, the performance gap between reasoning and non-reasoning models is especially pronounced on the BETTER dataset, which involves more complex guidelines and templates. Manual analysis revealed that constrained decoding is insufficient to enable non-reasoning models to produce good-quality templates. We hypothesize that the superior performance of reasoning models stems from their specialized training on long-form sequential dependencies, which facilitates more coherent, structured outputs in DocIE tasks, specifically in the BETTER dataset. Therefore, based on these results, the following analyses will focus exclusively on \textbf{reasoning models}.

\subsection{\textsc{ThinkTwice} as zero-shot}

\begin{table}[t]
    \centering
    \resizebox{\linewidth}{!}{
        \begin{tabular}{lc|ll}
            \toprule
            \textbf{Method} & \textbf{Selector} & \textbf{MultiMUC} & \textbf{BETTER}\\
            \midrule
            GPT \textsubscript{5.5} & \ding{55} &  19.84 & 27.95 \\
            \midrule
            
            Greedy\textsubscript{ Llama R1} & \ding{55} &  12.67 & 14.78  \\
            \multirow{3}{*}{\textsc{ThinkTwice}\textsubscript{ Llama R1}} & Majority & 13.83${\scriptstyle\pm}$\footnotesize{0.66} & 10.51${\scriptstyle\pm}$\footnotesize{2.30} \\
             & F1 Voting & \underline{14.41}${\scriptstyle\pm}$\footnotesize{0.33} & \underline{15.29}${\scriptstyle\pm}$\footnotesize{0.88} \\
            \rowcolor{lightgray!20} & (oracle)  & 31.77 & 34.08 \\
            \midrule
            Greedy\textsubscript{ Qwen3} & \ding{55} & 14.65 & 16.12 \\
            \multirow{3}{*}{\textsc{ThinkTwice}\textsubscript{ Qwen3}} & Majority & 16.18${\scriptstyle\pm}$\footnotesize{0.89} & 16.26${\scriptstyle\pm}$\footnotesize{0.82} \\
             & F1 Voting & \underline{\textbf{16.56}}${\scriptstyle\pm}$\footnotesize{0.37} & \underline{\textbf{19.45}}${\scriptstyle\pm}$\footnotesize{0.16} \\
            \rowcolor{lightgray!20} & (oracle) & 35.32 & 36.74 \\
            \bottomrule
        \end{tabular}
    }
    \caption{Zero-shot results on MultiMUC and BETTER. Greedy results are compared to two unsupervised selectors. \underline{Underline} and \textbf{bold} indicate the best result among greedy and the selectors per model and across all models, respectively.}
    \label{tab:selection_strategies}
\end{table}

Following the zero-shot experiments, we analyze the impact of \textsc{ThinkTwice} with different \textbf{unsupervised} selection strategies. Table~\ref{tab:selection_strategies} compares greedy decoding, our proposed \textsc{ThinkTwice} framework, and GPT\textsubscript{5.5}.

\paragraph{Does \textsc{ThinkTwice} improve zero-shot performance?}
Table~\ref{tab:selection_strategies} shows that sampling with an explicit selection strategy consistently outperforms greedy decoding. On MultiMUC, F1 Voting improves performance by roughly two F1 points across both backbones. Gains are even larger on the more challenging BETTER benchmark, where Majority Voting struggles with the increased template complexity, while F1 Voting yields consistent improvements over greedy decoding, exceeding three points for Qwen3. Additionally, F1 Voting exhibits lower variance in all datasets and models, indicating greater robustness than Majority Voting.

Overall, F1 Voting is the most effective unsupervised selection strategy, consistently outperforming greedy decoding across backbones and datasets.

\paragraph{How good are our current selection strategies?}
In addition to the results of the current selection strategies, Table~\ref{tab:selection_strategies} also reports the scores obtained using a perfect selector (oracle). Comparing the oracle results to GPT\textsubscript{5.5} single-sample results\footnote{Single-sample results are reported due to the lack of control over inference hyperparameters such as temperature and the resulting poor diversity (see Appendix~\ref{app:diversity}).}, we see that substantially smaller models can outperform proprietary state-of-the-art systems when multiple candidates are sampled, and an effective selector is used. However, the unsupervised selectors remain far below the oracle upper bound. This suggests that \textbf{there is still significant room for improvement through a more effective selection strategy}. To reduce this gap, we analyze a \textit{supervised} selection approach in the following sections.

\subsection{Supervised \textsc{ThinkTwice} }

All the results reported previously were obtained by an out-of-the-box LLM, without further training for the task. In this section, we explore the impact of supervised data along with our proposed rejection sampling method, and, the effectiveness of the supervised selector. We focus on MUC-4 for training and evaluation, and explore cross-lingual transfer in the following subsection.

\paragraph{Does \textsc{ThinkTwice} improve results in a supervised scenario?} Table~\ref{tab:supervised_english} reports the results for our supervised experiments on the MUC-4 dataset. As can be seen, similar conclusions are drawn compared to the zero-shot results: (1) \textsc{ThinkTwice} improves over the greedy baseline by a large margin; and, (2) F1 Voting performs consistently better than Majority Voting in all cases. Interestingly, when data is used to train the model, Majority Voting improvements are marginal or even detrimental, whereas F1 Voting results remain consistent.

\begin{table}[t]
\small
\centering
\begin{tabular}{lc|l}
\toprule
 \textbf{Method} &  \textbf{Selector} &  \textbf{MUC-4}\\
\midrule
 \textsc{TempGen}\textsubscript{ BART large} &  \ding{55} &  28.30 \\
 \textsc{GTT}\textsubscript{ BERT base} &  \ding{55} &  32.30 \\
 \textsc{IterX}\textsubscript{ T5-enc large} &  \ding{55} &  35.20 \\
\midrule
 Greedy\textsubscript{ Llama R1} &  \ding{55} &  28.52 \\
 \multirow{4}{*}{\textsc{ThinkTwice}\textsubscript{ Llama R1}} &  Majority &  28.41${\scriptscriptstyle\pm}$\tiny{1.49} \\
 &  F1 Voting &  36.56${\scriptscriptstyle\pm}$\tiny{0.22} \\
 &  Reward &  \underline{41.11}${\scriptscriptstyle\pm}$\tiny{0.48} \\
\rowcolor{lightgray!20} &  (oracle) &  56.60 \\
 \midrule
 Greedy\textsubscript{ Qwen3} &  \ding{55} &  35.97 \\
 \multirow{4}{*}{\textsc{ThinkTwice}\textsubscript{ Qwen3}} &  Majority &  35.67${\scriptscriptstyle\pm}$\tiny{0.59} \\
 &  F1 Voting &  40.77${\scriptscriptstyle\pm}$\tiny{0.53} \\
 &  Reward & \underline{\textbf{42.51}}${\scriptscriptstyle\pm}$\tiny{0.72} \\
\rowcolor{lightgray!20} &  (oracle) &  58.32 \\
\bottomrule
\end{tabular}
\caption{Supervised results for \textsc{ThinkTwice}, greedy baseline and the state-of-the-art on MUC-4.}
\label{tab:supervised_english}
\end{table}

While the performance gains over previous state-of-the-art methods are substantial, one might attribute this improvement to the superior inherent capabilities of the Llama R1 and Qwen3 compared to their predecessor models. Our results demonstrate that simply applying greedy decoding on modern models does not unlock their full potential, sometimes achieving performance below alternatives that leverage older and less capable LLMs.
In contrast, performance improves significantly when sampling is coupled with an effective selection strategy. Nevertheless, our approach still faces an important limitation, as it is constrained by the absence of gold-standard reasoning traces for supervised fine-tuning. Utilizing a rejection-sampling-based strategy, we successfully obtained gold-quality reasoning traces for only 67.77\% and 71.46\% of the training data for Llama R1 and Qwen3, respectively. A deeper analysis of the rejection-sampling fine-tuning is reported in Appendix \ref{app:rejection_sampling}. 

\paragraph{Is the supervised selector effective?}
Motivated by the intuition that classification is inherently easier than generation, we investigate the impact of training a selector on the data collected during the iterative rejection-sampling process. Table~\ref{tab:supervised_english} compares the supervised selector against the unsupervised approaches and the oracle upper bound (i.e., the maximum achievable score under perfect selection). The supervised selector consistently surpasses all unsupervised methods across both Llama and Qwen variants, with especially pronounced gains for the Llama R1 model. Incorporating the supervised selector yields further improvements for both models and establishes a new state-of-the-art for the task. Nevertheless, a substantial gap to the oracle remains. Although matching the oracle is likely unattainable in practice, the remaining margin suggests considerable room for improving upon our relatively simple supervised selector.

\begin{table*}[t]
\resizebox{\linewidth}{!}{
\begin{tabular}{lc|llllll|l}
\toprule
\textbf{Method} & \textbf{Selector} & \multicolumn{1}{l}{\textbf{English}} & \multicolumn{1}{l}{\textbf{Arabic}} & \multicolumn{1}{l}{\textbf{Farsi}} & \multicolumn{1}{l}{\textbf{Korean}} & \multicolumn{1}{l}{\textbf{Russian}} & \multicolumn{1}{l|}{\textbf{Chinese}} & \multicolumn{1}{l}{\textbf{Average}} \\
\midrule
GPT\textsubscript{ 5.5} & \ding{55} & 29.50 & 24.60 & 22.03 & 07.59 & 17.70 & 17.59 & 17.90 \\
\midrule
\citet{gantt-etal-2024-multimuc}\textsubscript{ Supervised} & \ding{55} & 35.20 & 21.46 & 20.66 & \textbf{23.91} & 23.77 & 21.93 & 22.35 \\
\midrule
\multirow{2}{*}{\textsc{ThinkTwice}\textsubscript{ Zero-shot}} & F1 Voting & 24.30${\scriptscriptstyle\pm}$\tiny{0.57} & 17.66${\scriptscriptstyle\pm}$\tiny{0.32} & 19.62${\scriptscriptstyle\pm}$\tiny{0.44} & 07.22${\scriptscriptstyle\pm}$\tiny{0.37} & 14.34${\scriptscriptstyle\pm}$\tiny{0.16} & 16.20${\scriptscriptstyle\pm}$\tiny{0.16} & 15.01${\scriptscriptstyle\pm}$\tiny{0.31} \\
 & Reward & 33.46${\scriptscriptstyle\pm}$\tiny{0.70} & 26.05${\scriptscriptstyle\pm}$\tiny{0.24} & 26.80${\scriptscriptstyle\pm}$\tiny{0.21} & 08.71${\scriptscriptstyle\pm}$\tiny{0.53} & 20.66${\scriptscriptstyle\pm}$\tiny{0.25} & 25.47${\scriptscriptstyle\pm}$\tiny{0.74} & 21.54${\scriptscriptstyle\pm}$\tiny{0.45} \\
 \midrule
\multirow{2}{*}{\textsc{ThinkTwice}\textsubscript{ English FT}} & F1 Voting & 40.77${\scriptscriptstyle\pm}$\tiny{0.53} & 22.66${\scriptscriptstyle\pm}$\tiny{0.44} & 23.61${\scriptscriptstyle\pm}$\tiny{0.77} & 08.73${\scriptscriptstyle\pm}$\tiny{0.75} & 20.53${\scriptscriptstyle\pm}$\tiny{1.18} & 22.15${\scriptscriptstyle\pm}$\tiny{0.92} & 19.54${\scriptscriptstyle\pm}$\tiny{0.85} \\
 & Reward & \textbf{42.51}${\scriptscriptstyle\pm}$\tiny{0.72} & \textbf{30.27}${\scriptscriptstyle\pm}$\tiny{0.49} & \textbf{30.16}${\scriptscriptstyle\pm}$\tiny{0.66} & 09.62${\scriptscriptstyle\pm}$\tiny{0.14} & \textbf{29.95}${\scriptscriptstyle\pm}$\tiny{1.23} & \textbf{29.75}${\scriptscriptstyle\pm}$\tiny{0.75} & \textbf{25.95}${\scriptscriptstyle\pm}$\tiny{0.74} \\
 \bottomrule
\end{tabular}
}
\caption{Cross-lingual transfer performance of \textsc{ThinkTwice}. The Qwen3-based LLM and reward model are trained exclusively on English data (MUC-4) and evaluated on multiple languages. Results are compared against state-of-the-art models trained on the target language \citep{gantt-etal-2024-multimuc}, as well as GPT\textsubscript{5.5} under a zero-shot setting. The averages are calculated without using the English scores.}
    \label{tab:crosslingual_analysis}
\end{table*}

\subsection{Cross-lingual transfer}

Finally, we evaluate the cross-lingual generalization of \textsc{ThinkTwice}. We first assess whether an LLM fine-tuned only on English data can transfer the learned knowledge to other languages when paired with the unsupervised selector. We then analyze the cross-lingual transferability of the learned reward selector. Lastly, we compare against the best previously reported multilingual MUC results from~\citet{gantt-etal-2024-multimuc}, which were obtained by fine-tuning language-specific models for the task.

\paragraph{Cross-lingual generalization of \textsc{ThinkTwice}.} Table~\ref{tab:crosslingual_analysis} presents the cross-lingual transfer results obtained with Qwen3 under both zero-shot and fine-tuned settings. In the zero-shot setting, performance drops substantially for languages other than English, with the largest degradation observed for Korean. Notably, GPT\textsubscript{5.5} exhibits a similar trend, further underscoring the persistent imbalance in multilingual capabilities across current language models. Fine-tuning on English data alone yields an average improvement of 4.53 F1 points across the other languages, demonstrating strong cross-lingual transfer despite the absence of target-language supervision. Although gains for Korean remain limited, most languages benefit consistently from English-only training. 

The introduction of a supervised selector leads to even larger improvements, providing 6.53 additional F1 points for the zero-shot model and 6.41 F1 points for the English fine-tuned model. Interestingly, combining both forms of supervision enables our models to outperform systems trained directly on language-specific data~\cite{gantt-etal-2024-multimuc} in four of the five languages, with Korean as the exception. Finally, compared to GPT\textsubscript{5.5}, we find that applying supervision either at the model level or at the selector level is already sufficient to surpass its performance, highlighting the critical role of supervised data in IE tasks.

\section{Conclusions}
We introduce \textsc{ThinkTwice}, a sampling and selection framework for document-level Information Extraction. The proposed method leverages the inherent output variability of decoder-only LLMs to recover higher-quality solutions than those obtained via greedy decoding. We further propose two complementary sample selection strategies: F1 Voting, which is fully unsupervised, and a supervised reward-based selector. Through an extensive experimental evaluation, we show that test-time scaling techniques can be effectively adapted to DocIE, surpassing greedy decoding and standard TTS baselines and achieving state-of-the-art results in supervised and cross-lingual settings.

We first demonstrate that reasoning-oriented LLMs consistently outperform non-reasoning models in zero-shot scenarios, with particularly strong gains on complex extraction tasks such as those in the BETTER dataset. Motivated by this observation, we conduct a comprehensive analysis across multiple datasets and languages, showing that the sampling and selection technique of \textsc{ThinkTwice} consistently outperforms greedy decoding. Additionally, the F1 Voting selector also surpasses the standard TTS approach, Majority Voting, in terms of performance and stability. 

In supervised experiments on MUC-4, we further showed that scaling model size alone is insufficient to produce substantial improvements when relying solely on greedy decoding. In contrast, combining stronger reasoning LLMs with \textsc{ThinkTwice} produces significant gains, highlighting the importance of effective test-time scaling strategies. We additionally demonstrated that the supervised selector can be trained entirely from automatically collected supervision, leading to further improvements beyond the unsupervised setting. Finally, training on English data alone was enough to transfer across languages: combining the fine-tuned Qwen3 reasoning model with the reward-based selector outperforms systems trained on target-language data in four of the five languages.

\section*{Limitations}
To apply our approach in a supervised setting, we must train reasoning-oriented LLMs, which requires high-quality reasoning traces. Although the proposed rejection-sampling procedure generated a substantial number of gold-quality trajectories, coverage remained incomplete, leading to noisy training signals. Future work could investigate reinforcement learning methods, such as GRPO, to train reasoning-oriented LLMs that perform more effectively on this task, potentially providing improvements complementary to those achieved through sampling and selection.

\textsc{ThinkTwice} also introduces additional computational overhead at inference time. Generating multiple candidate templates increases both latency and decoding cost relative to standard greedy decoding. Nevertheless, the number of generated samples (e.g., 64) is a tunable hyperparameter rather than a fixed constant, allowing a flexible trade-off between computational cost and performance.

Finally, despite the gains of F1 Voting and reward-based selection, a large gap to oracle performance remains, leaving room for better test-time scaling approaches for IE.

\section*{Acknowledgments}
This work has been partially supported by the European Union under Horizon Europe (Project LUMINOUS, grant number 101135724), the Spanish Ministry of Science, Innovation, and Universities (Project HumanAIze, grant number AIA2025-163322-C61) and the Basque Government (Research group funding IT2004-26). Mikel Zubillaga holds a PhD grant from the University of the Basque Country UPV/EHU (PIF24/04). The models were trained on the Leonardo supercomputer at CINECA under the EuroHPC Joint Undertaking, project EHPC-EXT-2024E01-042, and project EHPC-AIF-2026SC01-156. Some experiments were also run on the Calendula supercomputer at SCAYLE, supported by CLARIAH-ES.

\bibliography{custom}

\appendix
\section{Prompt}\label{app:prompt}

In order to define the Information Extraction task to the LLMs, we have created the prompt detailed below. For \textbf{Llama 3.3 70B} and \textbf{Qwen3 32B}, the prompt is structured as follows:

\paragraph{System Prompt:} You are an expert in information extraction, you need to extract the information of the document that is provided in \textit{\{language\}} as a template in JSON format. The guidelines for the dataset you need to extract are the followings: \textit{\{guidelines\}}
\paragraph{User Prompt:} Create the template for the next document: \textit{\{document\}}

In the case of \textbf{Llama R1 70B}, as system prompts are not recommended for it, the instructions are combined into a single user prompt.

\section{JSON Schemas}\label{app:json_schemas}
The inference process utilizes two distinct JSON schemas: one tailored for MUC-4 \citep{muc-1992-message} and MultiMUC \citep{gantt-etal-2024-multimuc}, and another specifically for the BETTER Granular \citep{mckinnon-rubino-2022-iarpa} dataset. While this overview outlines their primary structures, the complete schemas are documented in the supplementary materials.
\subsection{MUC-4 And MultiMUC}
This schema defines a list of templates for each document, where each template corresponds to a single incident. For every incident, the schema requires the explicit specification of an \textbf{incident\_type}, chosen from a fixed set of categories (kidnapping, attack, bombing, robbery, arson, or forced work stoppage).

In addition to the incident type, each template captures the main participants and elements involved in the event. These include the \textbf{PerpInd} (individual perpetrators) and \textbf{PerpOrg} (organizational perpetrators), represented as lists of names, as well as \textbf{Target}, which refers to inanimate objects or entities attacked. The schema also records the \textbf{Victim} slot, defined as people who were targeted or harmed, and the \textbf{Weapons} or devices used to carry out the incident. All participant-related fields are modeled as lists of strings to allow for multiple entities per incident.
\subsection{BETTER Granular}
The BETTER Granular dataset adopts a multi-template schema in which each document is annotated with a set of event templates. Each template corresponds to a specific event type, such as protests, terrorism, epidemics, disasters, displacement, or corruption, and is explicitly identified by a \textbf{template\_type} field. Templates comprise a structured collection of optional slots designed to capture salient aspects of an event, including participants, locations, temporal information, causes, and outcomes.

To illustrate this structure, consider the \textbf{Terrorplate} template, which represents terrorist incidents. Terrorplate captures the basic properties of an attack, including the type of attack, its completion status, and whether it was part of a coordinated or ongoing campaign. It records information about perpetrators, such as named individuals, organizations, or organizers, as well as attribution, blame, and claims of responsibility. The template also includes slots for targets, weapons, and casualties, covering both victims and perpetrators. Outcome fields allow the annotation of effects that occurred, were prevented, or were only potential. Finally, temporal and spatial fields specify when and where the incident took place.

Other templates in the BETTER schema follow the same design principles. Each focuses on a distinct event domain while employing a consistent slot-based structure. However, each template has its own slots that are needed in order to extract the concrete information of that event type.

\section{Guidelines}\label{app:guidelines}
This section provides a concise overview of the guidelines for the MUC-4 and MultiMUC datasets, as well as those for the BETTER Granular dataset. Full versions of all guidelines are available in Markdown format within the supplementary material.
\subsection{MUC-4 and MultiMUC}
The MUC-4 and MultiMUC guidelines begin with a concise overview of the MUC dataset. They then establish the criteria for relevant incidents through a set of specific inclusion and exclusion rules. Subsequently, the guidelines define the dataset slots and the required information for each. Finally, they specify the format for the output templates.
\subsection{BETTER Granular}
The BETTER Granular dataset guidelines begin by defining the dataset and its six constituent template types: Protestplate, Corruplate, Terrorplate, Epidemiplate, Disasterplate, and Displacementplate. Subsequently, the documentation provides a detailed description of each specific template type.

\begin{figure*}[t]
    \centering
    \begin{subfigure}[b]{0.495\linewidth}
        \includegraphics[width=\linewidth]{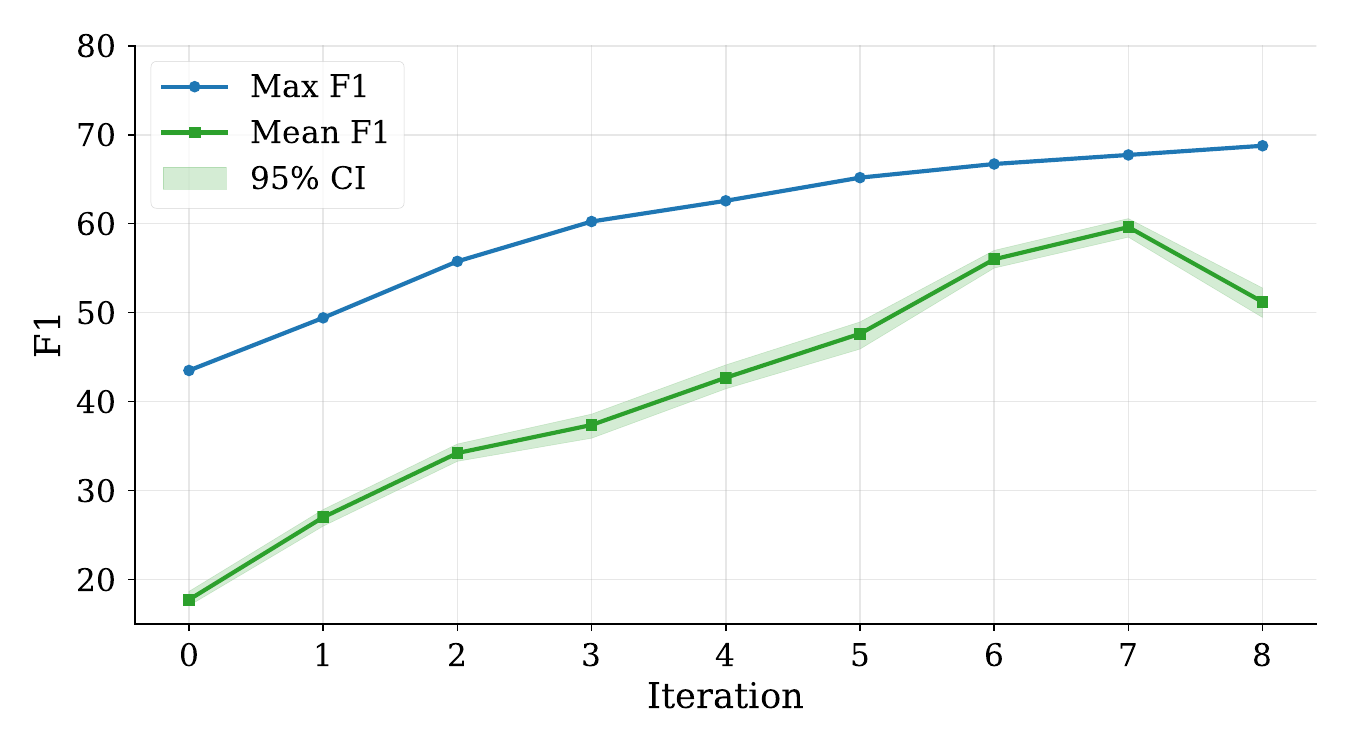}
        \caption{Llama R1 70B}
        \label{fig:scatter_ES}
    \end{subfigure}
    \begin{subfigure}[b]{0.495\linewidth}
        \includegraphics[width=\linewidth]{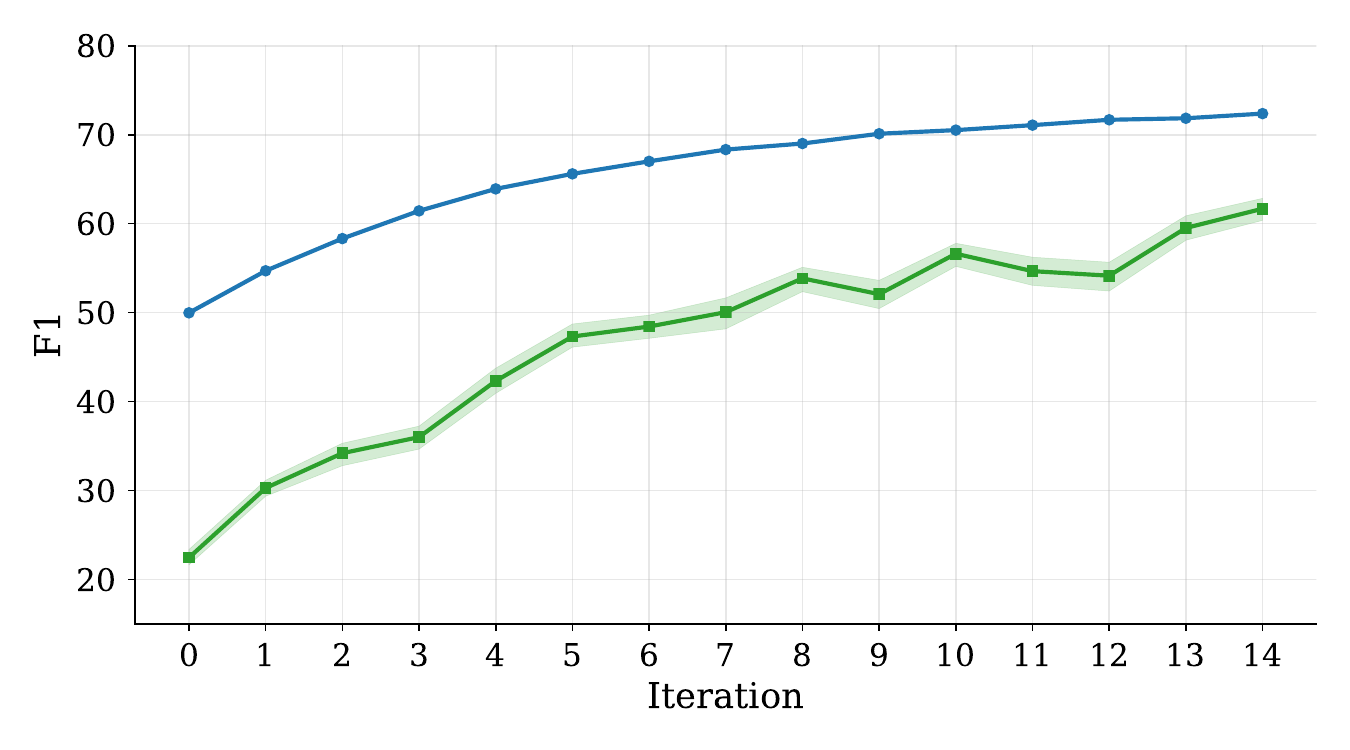}
        \caption{Qwen3 32B (Think)}
        \label{fig:scatter_EU}
    \end{subfigure}
    \caption{Maximum, mean, and 95\% confidence interval results in train split across different numbers of rejection sampling fine-tuning iterations.}
    \label{fig:rejection_sampling_analysis}
\end{figure*}

\section{Hyperparameters}\label{app:hyperparameters}
\begin{table}
\small
\centering
\begin{tabular}{l|cc|cc}
\toprule
 & \multicolumn{2}{c|}{Reasoning} & \multicolumn{2}{c}{Non-Reasoning} \\
Hyperparameter & \multicolumn{1}{c}{Llama} & \multicolumn{1}{c|}{Qwen} & \multicolumn{1}{c}{Llama} & \multicolumn{1}{c}{Qwen} \\
 \midrule
Temperature & 0.7 & 0.6 & 0.6 & 0.7 \\
Top-p & 1 & 0.95 & 1 & 0.8 \\
Top-k & -1 & 20 & -1 & 20 \\
Min-p & 0 & 0 & 0 & 0 \\
\bottomrule
\end{tabular}
\caption{Inference hyperparameters.}
\label{tab:inference_hyperparameters}
\end{table}

\begin{table}
\centering
\small
\begin{tabular}{l|cc}
\toprule
 Hyperparameter & Reasoning & Reward \\
 \midrule
Batch Size & 128 & 512 \\
Learning Rate & $5e^{-5}$& $2e^{-5}$\\
Epochs & 5 & 4 \\
Weight Decay & $5e^{-5}$& $5e^{-5}$\\
\bottomrule
\end{tabular}
\caption{Fine-tuning hyperparameters.}
\label{tab:finetuning_hyperparameters}
\end{table}

The \textsc{ThinkTwice} method requires several hyperparameters, which we categorize into three types: inference, fine-tuning, and rejection-sampling hyperparameters.

\paragraph{Inference Hyperparameters.} Since reasoning and non-reasoning models behave differently, they usually need to be sampled using different sampling settings. Table~\ref{tab:inference_hyperparameters} shows the inference hyperparameters that we have selected, which follow the default configurations recommended by the respective developers of each model. When sampling is required, \textbf{each model is sampled 64 times per prompt}. In those cases, to ensure reproducibility, we fix the random seed to $42$.

\paragraph{Fine-tuning Hyperparameters.} The fine-tuning hyperparameters are shown in Table \ref{tab:finetuning_hyperparameters}, in which we distinguish two types of fine-tuning: reasoning model fine-tuning and reward model fine-tuning. For the two types, the checkpoints are saved at each epoch, and the best-performing one on the development split is selected. As in the inference hyperparameters, a random seed of $42$ is used in order to ensure reproducibility. 

\paragraph{Rejection Sampling Hyperparameters.} We applied iterative rejection-sampling fine-tuning. Llama R1 converged after $8$ rounds, whereas Qwen3 required $14$ rounds. In both cases, the top $8$ templates were kept at each iteration. Once both processes had converged, we performed one additional round. In this final iteration, we used the top $4$ templates across the full dataset for Qwen3 and the top $1$ (i.e., the best-performing) template for Llama R1.

\section{Computational Infrastructure} \label{app:computational_infraestructure}
All experiments were conducted using NVIDIA A100 and H100 GPUs. The specific hardware allocation varied depending on the model scale and the nature of the computational task (fine-tuning and inference).

\paragraph{Fine-tuning Configuration.} For the \textbf{Llama R1 70B model}, we employed DeepSpeed ZeRO-3 \citep{rajbhandari2020zeromemoryoptimizationstraining} to manage memory sharding across 64 A100 GPUs. In contrast, the \textbf{Qwen3 32B} model was fine-tuned using 16 H100 GPUs utilizing the same ZeRO-3 configuration.

\paragraph{Inference Configuration.} Standard model inference was executed using Tensor Parallelism across 4 A100 GPUs to speed up generation. The only exception was the reward model, which, due to its smaller size, was deployed on a single A100 GPU.

\section{Output Diversity Analysis}
\label{app:diversity}

\begin{table}[h]
\small
\centering
\begin{tabular}{c|c}
\toprule
Model & F1 similarity \\
\midrule
GPT\textsubscript{5.5} & 71.76${\scriptscriptstyle\pm}$\tiny{12.65} \\
Llama R1 & 23.20${\scriptscriptstyle\pm}$\tiny{10.03} \\
Qwen3 & 26.72${\scriptscriptstyle\pm}$\tiny{11.93} \\
\bottomrule
\end{tabular}
\caption{Mean pairwise F1 similarity between N=64 sampled templates on the BETTER dataset for each model. Higher values indicate lower output diversity across samples.}
\label{tab:similarity_results}
\end{table}

In order to apply \textsc{ThinkTwice}, the model must produce sufficiently diverse predictions. While this is easily obtainable with standard inference hyperparameters (even for non-reasoning models), a low temperature, top-p or top-k value can make the model yield very similar outcomes.

When attempting to apply \textsc{ThinkTwice} to GPT\textsubscript{5.5}, we found that the API does not provide access to inference hyperparameters, preventing any direct control over sampling behavior. In addition, qualitative inspection of the generated outputs revealed very limited variability, with many predictions being nearly identical. Table~\ref{tab:similarity_results} quantifies this observation by reporting the mean pairwise F1 similarity between $N{=}64$ sampled templates on the BETTER dataset. GPT\textsubscript{5.5} shows a substantially higher similarity (71.76) compared to Llama R1 (23.2) and Qwen3 (26.72), indicating a markedly lower level of output diversity.

Given this limited variability and the lack of controllable sampling parameters, we restrict our experiments to open-source models, where inference conditions can be fully controlled and the diversity necessary for \textsc{ThinkTwice} can be reliably ensured.

\section{Analysis of Rejection Sampling fine-tuning}
\label{app:rejection_sampling}
We analyze rejection sampling fine-tuning by computing the maximum, mean, and 95\% confidence interval of the results at each iteration. To estimate the maximum performance, we select, for each document, the best-performing template generated by the model and use these to compute the final score. In contrast, calculating the exact mean and confidence intervals requires evaluating an intractable combinatorial space ($64^{1300}$). We therefore approximate them using a bootstrapping approach: for each document, we randomly sample one template to compute an overall score across the dataset, repeating this procedure 100 times. The mean and 95\% confidence intervals are then derived from these bootstrap samples.

The resulting values are presented in Figure~\ref{fig:rejection_sampling_analysis}, which illustrates the evolution of the rejection-sampling fine-tuned models across iterations. The maximum-performance curves show a clear convergence toward a horizontal asymptote in later iterations, stabilizing around 70 F1 points. In contrast, the mean score exhibits a consistent upward trend across most iterations. The confidence interval also changes over time, becoming wider in later iterations compared to earlier ones.

\section{Reasoning Model Budget}
\label{app:reasoning}

\begin{table*}[t]
\centering
\small
\begin{tabular}{l|ccccccc}
\toprule
 & \multicolumn{6}{c}{MultiMUC} & \multicolumn{1}{c}{BETTER} \\
\cmidrule(lr){2-7} \cmidrule(lr){8-8}
Model & English & Arabic & Farsi & Korean & Russian & Chinese & English \\
\midrule
Llama R1\textsubscript{ Zero-shot}& 364${\scriptscriptstyle\pm}$\tiny{146} & 368${\scriptscriptstyle\pm}$\tiny{127} & 379${\scriptscriptstyle\pm}$\tiny{125} & 364${\scriptscriptstyle\pm}$\tiny{132} & 560${\scriptscriptstyle\pm}$\tiny{221} & 485${\scriptscriptstyle\pm}$\tiny{140} & 447${\scriptscriptstyle\pm}$\tiny{138} \\
\midrule
Llama R1\textsubscript{ Fine-tuned}& 466${\scriptscriptstyle\pm}$\tiny{505} & 456${\scriptscriptstyle\pm}$\tiny{554} & 452${\scriptscriptstyle\pm}$\tiny{531} & 433${\scriptscriptstyle\pm}$\tiny{501} & 454${\scriptscriptstyle\pm}$\tiny{504} & 433${\scriptscriptstyle\pm}$\tiny{487} & -- \\
\midrule
Qwen3\textsubscript{ Zero-shot}& 792${\scriptscriptstyle\pm}$\tiny{578} & 749${\scriptscriptstyle\pm}$\tiny{519} & 750${\scriptscriptstyle\pm}$\tiny{505} & 826${\scriptscriptstyle\pm}$\tiny{563} & 809${\scriptscriptstyle\pm}$\tiny{561} & 709${\scriptscriptstyle\pm}$\tiny{466} & 701${\scriptscriptstyle\pm}$\tiny{364} \\
\midrule
Qwen3\textsubscript{ Fine-tuned}& 995${\scriptscriptstyle\pm}$\tiny{848} & 978${\scriptscriptstyle\pm}$\tiny{863} & 1005${\scriptscriptstyle\pm}$\tiny{868} & 1009${\scriptscriptstyle\pm}$\tiny{867} & 1013${\scriptscriptstyle\pm}$\tiny{855} & 994${\scriptscriptstyle\pm}$\tiny{873} & -- \\
\bottomrule
\end{tabular}
\caption{Mean and standard deviation of reasoning budget (in tokens) for each model and dataset, computed over all instances and traces.}
\label{tab:reasoning_budget}
\end{table*}

Table~\ref{tab:reasoning_budget} reports the mean and standard deviation of the reasoning budget used by each model. Several observations can be drawn from these results. First, the Llama R1 models consistently require fewer reasoning tokens than the Qwen3 models across all datasets and languages. Second, fine-tuned models generally use a larger reasoning budget than their zero-shot counterparts. Finally, the reasoning budget of the fine-tuned models is considerably more variable, as evidenced by their substantially larger standard deviations. This suggests that, although fine-tuning tends to increase the average amount of reasoning, it also leads to less consistent reasoning lengths across instances.

\section{Full Results} \label{app:full_results}
We present our experimental results below. The unsupervised results are divided into two tables: Table \ref{tab:BETTER_results} contains the results for the BETTER dataset, while Table \ref{tab:zero-shot_results} details the results for MUC-4 and MultiMUC. Finally, Table \ref{tab:supervised_results} summarizes all supervised results.

\begin{table}[h!]
\centering
\small
\begin{tabular}{lc|l}
\toprule
\textbf{Method} & \textbf{Selector} & \multicolumn{1}{l}{\textbf{BETTER}} \\
\midrule
GPT\textsubscript{ 5.5} & \ding{55} & \textbf{27.95} \\
\midrule
Greedy\textsubscript{ Llama 3.3} & \ding{55} & 3.20 \\
\multirow{3}{*}{\textsc{ThinkTwice}\textsubscript{ Llama 3.3}} & Majority & 1.72${\scriptscriptstyle\pm}$\tiny{0.42} \\
 & F1 & 1.60${\scriptscriptstyle\pm}$\tiny{0.22} \\
 \rowcolor{lightgray!20} & (oracle) & 5.71 \\
 \midrule
Greedy\textsubscript{ Llama R1} & \ding{55} & 14.78 \\
\multirow{3}{*}{\textsc{ThinkTwice}\textsubscript{ Llama R1}} & Majority & 10.51${\scriptscriptstyle\pm}$\tiny{2.30} \\
 & F1 & 15.29${\scriptscriptstyle\pm}$\tiny{0.88} \\
 \rowcolor{lightgray!20} & (oracle) & 34.08 \\
\midrule
Greedy\textsubscript{ Qwen no-think} & \ding{55} & 2.46 \\
\multirow{3}{*}{\textsc{ThinkTwice}\textsubscript{ Qwen no-think}} & Majority & 1.73${\scriptscriptstyle\pm}$\tiny{0.05} \\
 & F1 & 1.33${\scriptscriptstyle\pm}$\tiny{0.03} \\
 \rowcolor{lightgray!20} & (oracle) & 3.27 \\
\midrule
Greedy\textsubscript{ Qwen think} & \ding{55} & 16.12 \\
\multirow{3}{*}{\textsc{ThinkTwice}\textsubscript{ Qwen think}} & Majority & 16.26${\scriptscriptstyle\pm}$\tiny{0.82} \\
 & F1 & 19.45${\scriptscriptstyle\pm}$\tiny{0.16} \\
 \rowcolor{lightgray!20} & (oracle) & 36.74 \\
 \bottomrule
\end{tabular}
\caption{Zero-shot scenario results in the BETTER Granular dataset. We compare our approach using unsupervised selectors.}
\label{tab:BETTER_results}
\end{table}

\begin{table*}
\centering
\resizebox{\linewidth}{!}{
\begin{tabular}{lc|llllll|l}
\toprule
\textbf{Method} & \textbf{Selector} & \multicolumn{1}{l}{\textbf{English}} & \multicolumn{1}{l}{\textbf{Arabic}} & \multicolumn{1}{l}{\textbf{Farsi}} & \multicolumn{1}{l}{\textbf{Korean}} & \multicolumn{1}{l}{\textbf{Russian}} & \multicolumn{1}{l|}{\textbf{Chinese}} & \multicolumn{1}{l}{\textbf{Average}} \\
\midrule
Reference & \ding{55} & 35.2 & 21.46 & 20.66 & \textbf{23.91} & 23.77 & 21.93 & 24.49 \\
\midrule
Greedy\textsubscript{ Llama R1} & \ding{55} & 28.52 & 3.57 & 1.30 & 3.07 & 0.51 & 4.21 & 6.86 \\
\multirow{4}{*}{\textsc{ThinkTwice}\textsubscript{ Llama R1}} & Majority & 28.41${\scriptscriptstyle\pm}$\tiny{1.49} & 2.69${\scriptscriptstyle\pm}$\tiny{0.50} & 0.53${\scriptscriptstyle\pm}$\tiny{0.57} & 2.29${\scriptscriptstyle\pm}$\tiny{0.15} & 0.06${\scriptscriptstyle\pm}$\tiny{0.11} & 4.72${\scriptscriptstyle\pm}$\tiny{0.84} & 6.45${\scriptscriptstyle\pm}$\tiny{0.77} \\
 & F1 Voting & 36.56${\scriptscriptstyle\pm}$\tiny{0.22} & 3.79${\scriptscriptstyle\pm}$\tiny{0.30} & 1.43${\scriptscriptstyle\pm}$\tiny{0.39} & 3.88${\scriptscriptstyle\pm}$\tiny{0.31} & 0.24${\scriptscriptstyle\pm}$\tiny{0.10} & 5.34${\scriptscriptstyle\pm}$\tiny{0.46} & 8.54${\scriptscriptstyle\pm}$\tiny{0.32} \\
 & Reward & 41.11${\scriptscriptstyle\pm}$\tiny{0.48} & 5.65${\scriptscriptstyle\pm}$\tiny{0.15} & 2.43${\scriptscriptstyle\pm}$\tiny{0.34} & 5.08${\scriptscriptstyle\pm}$\tiny{0.21} & 1.36${\scriptscriptstyle\pm}$\tiny{0.21} & 10.98${\scriptscriptstyle\pm}$\tiny{0.27} & 11.10${\scriptscriptstyle\pm}$\tiny{0.30} \\
  \rowcolor{lightgray!20} & (oracle) & 56.60 & 12.79 & 4.66 & 11.82 & 2.96 & 19.44 & 18.04 \\
  \midrule
Greedy\textsubscript{ Qwen think} & \ding{55} & 35.97 & 17.60 & 16.17 & 7.79 & 15.33 & 16.14 & 18.17 \\
\multirow{4}{*}{\textsc{ThinkTwice}\textsubscript{ Qwen think}} & Majority & 35.67${\scriptscriptstyle\pm}$\tiny{0.59} & 18.20${\scriptscriptstyle\pm}$\tiny{0.25} & 17.13${\scriptscriptstyle\pm}$\tiny{0.87} & 5.78${\scriptscriptstyle\pm}$\tiny{0.70} & 16.11${\scriptscriptstyle\pm}$\tiny{1.60} & 15.26${\scriptscriptstyle\pm}$\tiny{0.28} & 18.02${\scriptscriptstyle\pm}$\tiny{0.85} \\
 & F1 Voting & 40.77${\scriptscriptstyle\pm}$\tiny{0.53} & 22.66${\scriptscriptstyle\pm}$\tiny{0.44} & 23.61${\scriptscriptstyle\pm}$\tiny{0.77} & 8.73${\scriptscriptstyle\pm}$\tiny{0.75} & 20.53${\scriptscriptstyle\pm}$\tiny{1.18} & 22.15${\scriptscriptstyle\pm}$\tiny{0.92} & 23.07${\scriptscriptstyle\pm}$\tiny{0.80} \\
 & Reward & \textbf{42.51}${\scriptscriptstyle\pm}$\tiny{0.72} & \textbf{30.27}${\scriptscriptstyle\pm}$\tiny{0.49} & \textbf{30.16}${\scriptscriptstyle\pm}$\tiny{0.66} & 9.62${\scriptscriptstyle\pm}$\tiny{0.14} & \textbf{29.95}${\scriptscriptstyle\pm}$\tiny{1.23} & \textbf{29.75}${\scriptscriptstyle\pm}$\tiny{0.75} & \textbf{28.71}${\scriptscriptstyle\pm}$\tiny{0.74} \\
  \rowcolor{lightgray!20} & (oracle) & 58.32 & 45.52 & 46.74 & 22.63 & 45.14 & 46.73 & 44.18 \\
  \bottomrule
\end{tabular}
}
\caption{Supervised and cross-lingual results on MUC-4 (English) and MultiMUC comparing greedy decoding against different selection strategies in reasoning fine-tuned models on English data. We also report the results of the state-of-the-art method, IterX\textsubscript{ T5-enc large} \citep{chen-etal-2023-iterative} fine-tuned using both English and target-language data for training (Reference) reported in the work of \citep{gantt-etal-2024-multimuc}.}
\label{tab:supervised_results}
\end{table*}

\begin{table*}
\centering
\resizebox{\linewidth}{!}{
\begin{tabular}{lc|llllll|l}
\toprule
\textbf{Method} & \textbf{Selector} & \multicolumn{1}{l}{\textbf{English}} & \multicolumn{1}{l}{\textbf{Arabic}} & \multicolumn{1}{l}{\textbf{Farsi}} & \multicolumn{1}{l}{\textbf{Korean}} & \multicolumn{1}{l}{\textbf{Russian}} & \multicolumn{1}{l|}{\textbf{Chinese}} & \multicolumn{1}{l}{\textbf{Average}} \\
\midrule

GPT\textsubscript{ 5.5} & \ding{55} & 29.50 & 24.60 & 22.03 & 7.59 & 17.70 & 17.59 & 19.84 \\
\midrule
Greedy\textsubscript{ Llama 3.3} & \ding{55} & 18.27 & 14.52 & 14.57 & 5.55 & 10.84 & 11.02 & 12.46 \\
\multirow{4}{*}{\textsc{ThinkTwice}\textsubscript{ Llama 3.3}} & Majority & 18.69${\scriptscriptstyle\pm}$\tiny{0.19} & 14.62${\scriptscriptstyle\pm}$\tiny{0.63} & 15.09${\scriptscriptstyle\pm}$\tiny{0.18} & 5.69${\scriptscriptstyle\pm}$\tiny{0.12} & 11.27${\scriptscriptstyle\pm}$\tiny{0.43} & 11.68${\scriptscriptstyle\pm}$\tiny{0.27} & 12.84${\scriptscriptstyle\pm}$\tiny{0.35} \\

 & F1 Voting & 18.60${\scriptscriptstyle\pm}$\tiny{0.14} & 14.80${\scriptscriptstyle\pm}$\tiny{0.11} & 15.02${\scriptscriptstyle\pm}$\tiny{0.06} & 5.79${\scriptscriptstyle\pm}$\tiny{0.17} & 10.99${\scriptscriptstyle\pm}$\tiny{0.18} & 11.62${\scriptscriptstyle\pm}$\tiny{0.18} & 12.80${\scriptscriptstyle\pm}$\tiny{0.15} \\

 & Reward & 22.84${\scriptscriptstyle\pm}$\tiny{0.36} & 18.83${\scriptscriptstyle\pm}$\tiny{0.13} & 18.60${\scriptscriptstyle\pm}$\tiny{0.62} & 7.26${\scriptscriptstyle\pm}$\tiny{0.20} & 15.04${\scriptscriptstyle\pm}$\tiny{0.28} & 17.01${\scriptscriptstyle\pm}$\tiny{0.03} & 16.60${\scriptscriptstyle\pm}$\tiny{0.33} \\

 \rowcolor{lightgray!20} & (oracle) & 26.76 & 25.71 & 26.07 & 10.87 & 19.37 & 20.51 & 21.55 \\
 \midrule
Greedy\textsubscript{ Llama R1} & \ding{55} & 18.68 & 15.11 & 12.55 & 5.53 & 11.85 & 12.27 & 12.67 \\
\multirow{4}{*}{\textsc{ThinkTwice}\textsubscript{ Llama R1}} & Majority & 21.52${\scriptscriptstyle\pm}$\tiny{0.76} & 16.23${\scriptscriptstyle\pm}$\tiny{0.96} & 14.42${\scriptscriptstyle\pm}$\tiny{0.62} & 5.74${\scriptscriptstyle\pm}$\tiny{0.28} & 12.31${\scriptscriptstyle\pm}$\tiny{0.75} & 12.78${\scriptscriptstyle\pm}$\tiny{0.29} & 13.83${\scriptscriptstyle\pm}$\tiny{0.66} \\
 & F1 Voting & 21.69${\scriptscriptstyle\pm}$\tiny{0.15} & 16.81${\scriptscriptstyle\pm}$\tiny{0.50} & 16.26${\scriptscriptstyle\pm}$\tiny{0.38} & 6.18${\scriptscriptstyle\pm}$\tiny{0.26} & 12.63${\scriptscriptstyle\pm}$\tiny{0.25} & 12.87${\scriptscriptstyle\pm}$\tiny{0.29} & 14.41${\scriptscriptstyle\pm}$\tiny{0.33} \\
 & Reward & 31.43${\scriptscriptstyle\pm}$\tiny{0.80} & 24.58${\scriptscriptstyle\pm}$\tiny{0.41} & 21.15${\scriptscriptstyle\pm}$\tiny{0.38} & 8.02${\scriptscriptstyle\pm}$\tiny{0.60} & 18.32${\scriptscriptstyle\pm}$\tiny{0.25} & 19.68${\scriptscriptstyle\pm}$\tiny{0.77} & 20.53${\scriptscriptstyle\pm}$\tiny{0.57} \\
 \rowcolor{lightgray!20} & (oracle) & 42.32 & 36.06 & 36.78 & 16.87 & 29.89 & 28.72 & 31.77 \\
 \midrule
Greedy\textsubscript{ Qwen no-think} & \ding{55} & 17.71 & 14.30 & 13.74 & 5.26 & 10.54 & 12.43 & 12.33 \\
\multirow{4}{*}{\textsc{ThinkTwice}\textsubscript{ Qwen no-think}} & Majority & 17.99${\scriptscriptstyle\pm}$\tiny{0.17} & 14.83${\scriptscriptstyle\pm}$\tiny{0.26} & 14.22${\scriptscriptstyle\pm}$\tiny{0.28} & 5.78${\scriptscriptstyle\pm}$\tiny{0.43} & 10.85${\scriptscriptstyle\pm}$\tiny{0.09} & 12.99${\scriptscriptstyle\pm}$\tiny{0.46} & 12.78${\scriptscriptstyle\pm}$\tiny{0.31} \\
 & F1 Voting & 18.22${\scriptscriptstyle\pm}$\tiny{0.10} & 15.30${\scriptscriptstyle\pm}$\tiny{0.05} & 14.49${\scriptscriptstyle\pm}$\tiny{0.08} & 5.71${\scriptscriptstyle\pm}$\tiny{0.26} & 11.01${\scriptscriptstyle\pm}$\tiny{0.08} & 13.79${\scriptscriptstyle\pm}$\tiny{0.46} & 13.09${\scriptscriptstyle\pm}$\tiny{0.22} \\
 & Reward & 24.06${\scriptscriptstyle\pm}$\tiny{0.25} & 17.27${\scriptscriptstyle\pm}$\tiny{0.07} & 16.61${\scriptscriptstyle\pm}$\tiny{0.31} & 7.14${\scriptscriptstyle\pm}$\tiny{0.27} & 13.94${\scriptscriptstyle\pm}$\tiny{0.15} & 17.74${\scriptscriptstyle\pm}$\tiny{0.26} & 16.13${\scriptscriptstyle\pm}$\tiny{0.23} \\
 \rowcolor{lightgray!20} & (oracle) & 26.92 & 22.05 & 21.79 & 9.62 & 18.11 & 21.46 & 19.99 \\
 \midrule
Greedy\textsubscript{ Qwen think} & \ding{55} & 22.99 & 14.28 & 16.13 & 6.88 & 11.68 & 15.92 & 14.65 \\
\multirow{4}{*}{\textsc{ThinkTwice}\textsubscript{ Qwen think}} & Majority & 25.80${\scriptscriptstyle\pm}$\tiny{0.87} & 15.81${\scriptscriptstyle\pm}$\tiny{0.38} & 17.47${\scriptscriptstyle\pm}$\tiny{1.24} & 6.93${\scriptscriptstyle\pm}$\tiny{0.88} & 14.66${\scriptscriptstyle\pm}$\tiny{0.75} & 16.43${\scriptscriptstyle\pm}$\tiny{0.98} & 16.18${\scriptscriptstyle\pm}$\tiny{0.89} \\
 & F1 Voting & 24.30${\scriptscriptstyle\pm}$\tiny{0.57} & 17.66${\scriptscriptstyle\pm}$\tiny{0.32} & 19.62${\scriptscriptstyle\pm}$\tiny{0.44} & 7.22${\scriptscriptstyle\pm}$\tiny{0.37} & 14.34${\scriptscriptstyle\pm}$\tiny{0.16} & 16.20${\scriptscriptstyle\pm}$\tiny{0.16} & 16.56${\scriptscriptstyle\pm}$\tiny{0.37} \\
 & Reward & \textbf{33.46}${\scriptscriptstyle\pm}$\tiny{0.70} & \textbf{26.05}${\scriptscriptstyle\pm}$\tiny{0.24} & \textbf{26.80}${\scriptscriptstyle\pm}$\tiny{0.21} & \textbf{8.71}${\scriptscriptstyle\pm}$\tiny{0.53} & \textbf{20.66}${\scriptscriptstyle\pm}$\tiny{0.25} & \textbf{25.47}${\scriptscriptstyle\pm}$\tiny{0.74} & \textbf{23.53}${\scriptscriptstyle\pm}$\tiny{0.50} \\
 \rowcolor{lightgray!20} & (oracle) & 46.48 & 39.15 & 39.00 & 17.94 & 31.21 & 38.13 & 35.32 \\
 \bottomrule
\end{tabular}
}
\caption{Zero-shot scenario results. We compare our approach using unsupervised selectors (F1 and Majority Voting) and supervised selectors (Reward trained in English data) in MUC-4 (English) and MultiMUC.}
\label{tab:zero-shot_results}
\end{table*}

\end{document}